\documentclass[10pt,twocolumn,letterpaper]{article}

\usepackage{cvpr}              
\usepackage{graphicx}
\usepackage{amsmath}
\usepackage{amssymb}
\usepackage{booktabs}

\usepackage{comment}
\usepackage{multirow}
\usepackage{bm}
\usepackage{enumitem}
\usepackage{listings}
\usepackage{dblfloatfix}
 
\usepackage[T1]{fontenc}

\newcommand{\deva}[1]{{\color{red}[Deva: #1]}}

\newcommand{\R}{\mathbb{R}}

\usepackage{xcolor}

\definecolor{codegreen}{rgb}{0,0.6,0}
\definecolor{codegray}{rgb}{0.5,0.5,0.5}
\definecolor{codepurple}{rgb}{0.58,0,0.82}
\definecolor{backcolour}{rgb}{0.95,0.95,0.92}

\lstdefinestyle{mystyle}{
    backgroundcolor=\color{backcolour},   
    commentstyle=\color{codegreen},
    keywordstyle=\color{magenta},
    numberstyle=\tiny\color{codegray},
    stringstyle=\color{codepurple},
    basicstyle=\ttfamily\footnotesize,
    breakatwhitespace=false,         
    breaklines=true,                 
    captionpos=b,                    
    keepspaces=true,                 
    numbers=right,                    
    numbersep=5pt,                  
    showspaces=false,                
    showstringspaces=false,
    showtabs=false,                  
    tabsize=2
}

\usepackage[pagebackref,breaklinks,colorlinks]{hyperref}

\usepackage[capitalize]{cleveref}
\crefname{section}{Sec.}{Secs.}
\Crefname{section}{Section}{Sections}
\Crefname{table}{Table}{Tables}
\crefname{table}{Tab.}{Tabs.}

\def\cvprPaperID{4005} 
\def\confName{CVPR}
\def\confYear{2023}

\begin{document}

\title{Soft Augmentation for Image Classification}

\author{Yang Liu, Shen Yan, Laura Leal-Taixé, James Hays, Deva Ramanan\\
Argo AI\\
{\tt\small youngleoel@gmail.com, shenyan@google.com, leal.taixe@tum.de, hays@gatech.edu, deva@cs.cmu.edu}
}

\maketitle

\begin{abstract} 
  Modern neural networks are over-parameterized and thus rely on strong regularization such as data augmentation and weight decay to reduce overfitting and improve generalization. The dominant form of data augmentation applies invariant transforms, where the learning target of a sample is invariant to the transform applied to that sample. We draw inspiration from human visual classification studies and propose generalizing augmentation with invariant transforms to soft augmentation where the \textbf{learning target softens non-linearly as a function of the degree of the transform} applied to the sample: e.g., more aggressive image crop augmentations produce less confident learning targets. We demonstrate that soft targets allow for more aggressive data augmentation, offer more robust performance boosts, work with other augmentation policies, and interestingly, produce better calibrated models (since they are trained to be less confident on aggressively cropped/occluded examples).
  Combined with existing aggressive augmentation strategies, soft targets 1) {\bf double} the top-1 accuracy boost across Cifar-10, Cifar-100, ImageNet-1K, and ImageNet-V2, 2) improve model occlusion performance by up to \bm{$4\times$}, and 3) {\bf half} the expected calibration error (ECE). 
  Finally, we show that soft augmentation generalizes to self-supervised classification tasks. Code available at \url{https://github.com/youngleox/soft_augmentation}
  
  
  \begin{figure*}[ht]
    \centering
    \begin{tabular}{c}
          \includegraphics[width=0.85\linewidth]{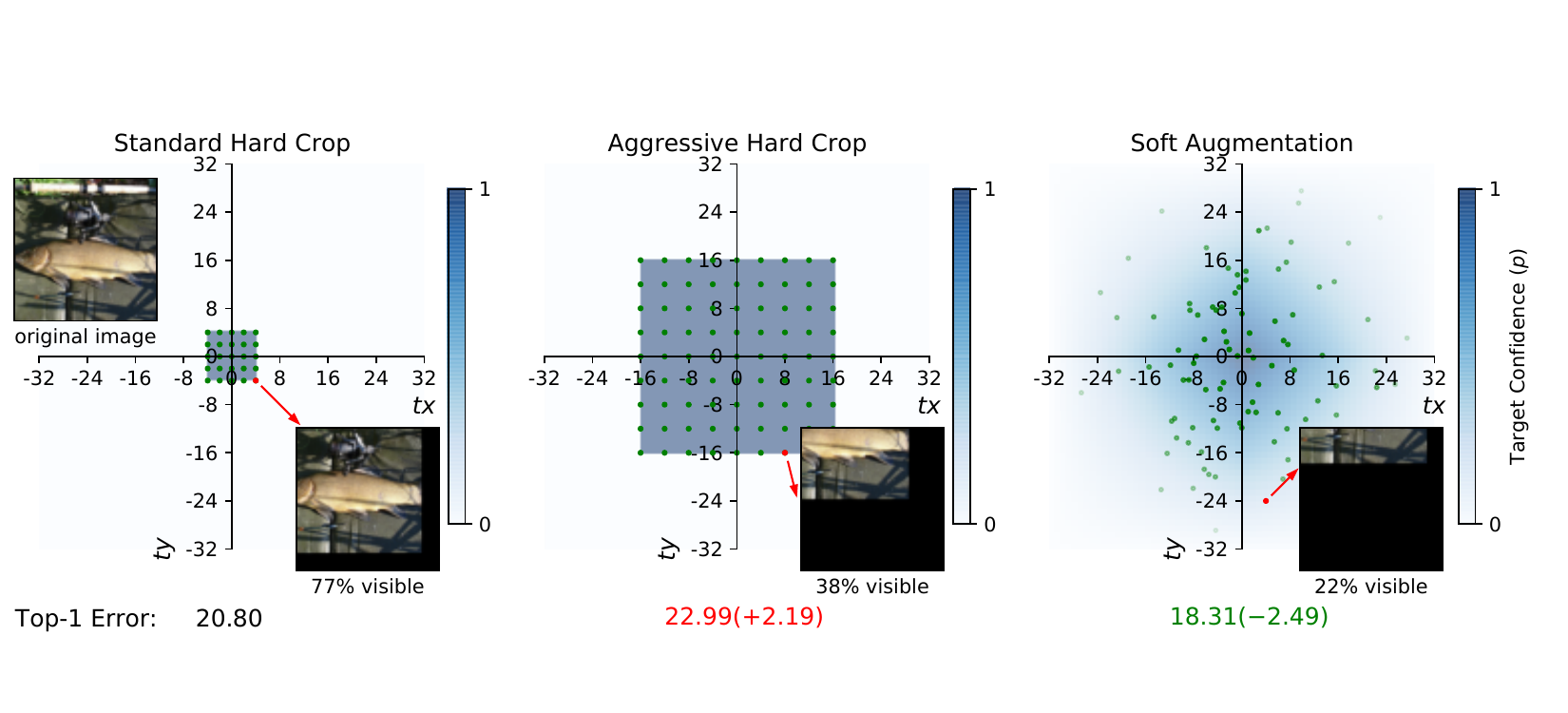} \\
    \end{tabular}
    \vspace{-8mm}
    \caption{
    Traditional augmentation encourages invariance by requiring  augmented samples to produce the same target label; we visualize the translational offset range $(tx,ty)$ of Standard Hard Crop augmentations for $32 \times 32$ images from Cifar-100 on the {\bf left}, reporting the top-1 error of a baseline ResNet-18. Naively increasing the augmentation range without reducing target confidence increases error ({\bf middle}), but softening the target label by reducing the target confidence for extreme augmentations reduces the error ({\bf right}), allowing for training with even more aggressive augmentations that may even produce \emph{blank images}. Our work also shows that soft augmentations produce models that are more robust to occlusions (since they encounter larger occlusions during training) and models that are better calibrated (since they are trained to be less-confident on such occluded examples).
    }
    \label{fig:hard_vs_soft}
    \vspace{-3mm}
\end{figure*}
\end{abstract}

\section{Introduction}


Deep neural networks have enjoyed great success in the past decade in domains such as visual understanding~\cite{googlenet}, natural language processing~\cite{brown2020gpt}, and protein structure prediction~\cite{senior2020alphafold}. However, modern deep learning models are often over-parameterized and prone to overfitting. In addition to designing models with better inductive biases, strong regularization techniques such as weight decay and data augmentation are often necessary for neural networks to achieve ideal performance. Data augmentation is often a computationally cheap and effective way to regularize models and mitigate overfitting. The dominant form of data augmentation modifies training samples with invariant transforms -- transformations of the data where it is assumed that the identity of the sample is \emph{invariant} to the transforms.

Indeed, the notion of visual invariance is supported by evidence found from biological visual systems~\cite{zhu2019robustness}. The robustness of human visual recognition has long been documented and inspired many learning methods including data augmentation and architectural improvement~\cite{wyatte2012limits,stn}. This paper focuses on the counterpart of human visual robustness, namely \textbf{how our vision fails}. 
Instead of maintaining perfect invariance, human visual confidence degrades \textbf{non-linearly} as a function of the degree of transforms such as occlusion, likely as a result of information loss~\cite{tang2018recurrent}
. We propose modeling the transform-induced information loss for learned image classifiers and summarize the contributions as follows:
\begin{itemize}[leftmargin=*]
\item We propose Soft Augmentation as a generalization of data augmentation with invariant transforms. With Soft Augmentation, the learning target of a transformed training sample \emph{softens}. We empirically compare several softening strategies and prescribe a robust non-linear softening formula.

\item With a frozen softening strategy, we show that replacing standard crop augmentation with soft crop augmentation allows for more aggressive augmentation, and \textbf{doubles} the top-1 accuracy boost of RandAugment~\cite{cubuk2020randaugment} across Cifar-10, Cifar-100, ImageNet-1K, and ImageNet-V2.

\item Soft Augmentation improves model occlusion robustness by achieving up to more than \bm{$\>4\times$} Top-1 accuracy boost on heavily occluded images.

\item Combined with TrivialAugment~\cite{muller2021trivialaugment}, Soft Augmentation further reduces top-1 error and improves model calibration by reducing expected calibration error by more than \textbf{half}, outperforming 5-ensemble methods~\cite{ensemble}.

\item In addition to supervised image classification models, Soft Augmentation also boosts the performance of self-supervised models, demonstrating its generalizability.
\end{itemize}

\section{Related Work} \label{sec:related}

\subsection{Neural Networks for Vision}

Since the seminal work from Krizhevsky \etal~\cite{alexnet}, neural networks have been the dominant class of high performing visual classifiers. Convolutional Neural Networks (CNNs) are a popular family of high performing neural models which borrows a simple idea of spatially local computations from biological vision~\cite{hubel1959receptive, fukushima1982neocognitron, lecun2015deep}. With the help of architectural improvements~\cite{resnet}, auxiliary loss~\cite{googlenet}, and improved computational power~\cite{lr_scaling}, deeper, larger, and more efficient neural nets have been developed in the past decade. 

\subsection{Data Augmentation}

Data augmentation has been an essential regularizer for high performing neural networks in many domains including visual recognition. While many other regularization techniques such as weight decay~\cite{nero} and batch normalization~\cite{brock2021nfnet} are shown to be optional, we are aware of no competitive vision models that omit data augmentation.

Accompanying the influential AlexNet model, Krizhevsky \etal~\cite{alexnet} proposed horizontal flipping and random cropping transforms which became the backbone of image data augmentation. Since the repertoire of invariant transformations has grown significantly in the past decade~\cite{googlenet}, choosing which subset to use and then finding the optimal hyperparameters for each transform has become computationally burdensome. This sparked a line of research~\cite{cubuk2019autoaugment,faa} which investigates optimal policies for data augmentation such as RandAugment~\cite{cubuk2020randaugment} and TrivialAugment~\cite{muller2021trivialaugment}.

\subsection{Learning from Soft Targets}

While minimizing the cross entropy loss between model logits and hard one-hot targets remains the go-to recipe for supervised classification training, learning with \emph{soft} targets has emerged in many lines of research. Label Smoothing~\cite{label_smoothing, label_smoothing_when} is a straightforward method which applies a fixed smoothing (softening) factor $\alpha$ to the hard one-hot classification target. The motivation is that label smoothing prevents the model from becoming over-confident. M{\"u}ller \etal~\cite{label_smoothing_when} shows that label smoothing is related to knowledge distillation~\cite{hinton2015distilling}, where a student model learns the soft distribution of a (typically) larger teacher model.

A related line of research~\cite{zhang2017mixup,yun2019cutmix} focuses on regularizing how a model interpolates between samples by linearly mixing two or more samples and linearly softening the resulting learning targets. Mixing can be in the form of per-pixel blending~\cite{zhang2017mixup} or patch-level recombination~\cite{yun2019cutmix}.

\subsection{Robustness of Human Vision}

Human visual classification is known to be robust against perturbations such as occlusion. In computer vision research, the robustness of human vision is often regarded as the gold standard for designing computer vision models~\cite{zhu2019robustness, lowe1999invariant_features}. These findings indeed inspire development of robust vision models, such as compositional, recurrent, and occlusion aware models~\cite{wang2020robust,wyatte2012limits,kortylewski2020compositional}. In addition to specialty models, much of the idea of using invariant transforms to augment training samples come from the intuition and observation that human vision are robust against these transforms such as object translation, scaling, occlusion, photometric distortions, etc.

Recent studies such as Tang \etal~\cite{tang2018recurrent} indeed confirm the robustness of human visual recognition against mild to moderate perturbations. In a 5-class visual classification task, human subjects maintain high accuracy when up to approximately half of an object is occluded. However, the more interesting observation is that human performance starts to degenerate rapidly as occlusion increases and falls to chance level when the object is fully occluded (see Figure \ref{fig:soften_methods} right $k=2,3,4$ for qualitative curves). 
\section{Soft Augmentation}\label{sec:soft_augmentation}

\begin{figure*}[ht]
    \vspace{-4mm}
    \centering
    \begin{tabular}{c c}
          \includegraphics[width=0.42\linewidth]{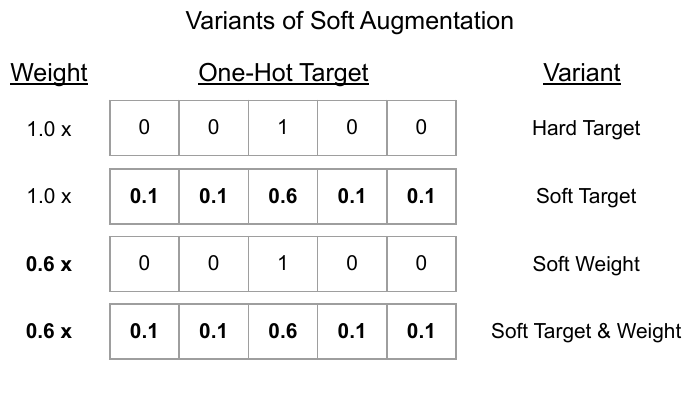}
          & 
          \includegraphics[width=0.26\linewidth]{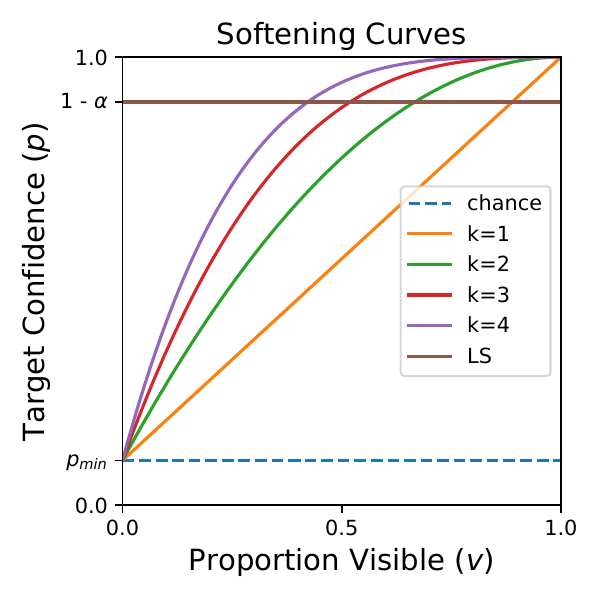}
          \\
    \end{tabular}
    \caption{Variants of Soft Augmentation as prescribed by Equations \ref{eq:soft_augment}  (\textit{Soft Target}), \ref{eq:soft_augment_w} (\textit{Soft Weight}), \ref{eq:soft_augment_tw} (\textit{Soft Target \& Weight}) with example target confidence $p=0.6$ (left). Soft Augmentation applies non-linear ($k=2,3,4,...)$ softening to learning targets based on \textbf{the specific degree of occlusion of a cropped image} (Equation \ref{eq:soften_curve}), which qualitatively captures the degradation of human visual recognition under occlusion~\cite{tang2018recurrent}. Label Smoothing applies a \textbf{fixed} softening factor $\alpha$ to the one-hot classification target. }
    \label{fig:soften_methods}
    \vspace{-3mm}
\end{figure*}

In a typical supervised image classification setting, each training image $x_i$ has a ground truth learning target $y_i$ associated to it thus forming tuples: 
\begin{equation}\label{eq:learning_sample}
    (x_i,y_i), 
\end{equation}
where $x_i \in \R^{C \times W \times H}$ denotes the image and $y_i \in [0,1]^{N}$ denotes a $N$-dimensional one-hot vector representing the target label (Figure \ref{fig:soften_methods} left, ``Hard Target'').
As modern neural models have the capacity to memorize even large datasets~\cite{memorization}, data augmentation mitigates the issue by hallucinating data points through transformations of existing training samples.

\textbf{(Hard) data augmentation} 
relies on the key underlying assumption that the augmented variant of $x_i$ should maintain the original target label $y_i$:
\begin{align}\label{eq:hard_augment}
    (x_i,y_i) \Rightarrow \left(t_{\phi \sim S}(x_i), y_i \right), \qquad \text{\bf Hard Augmentation}
\end{align}
where $t_{\phi \sim S }(x_i)$ denotes the image transform applied to sample $x_i$, $\phi$ is a random sample from the fixed transform range $S$. Examples of image transforms include translation, rotation, crop, noise injection, etc. 
As shown by Tang \etal~\cite{tang2018recurrent}, transforms of $x_i$ such as occlusion are approximately perceptually invariant only 
when $\phi$ is mild. Hence $S$ often has to be carefully tuned in practice, since naively increasing it can lead to degraded performance (Figure \ref{fig:hard_vs_soft}). In the extreme case of 100\% occlusion, total information loss occurs, making it detrimental for learning.

\textbf{Label Smoothing} applies a smoothing function $g$ to the target label $y_i$ parameterized by a handcrafted, fixed smoothing factor $\alpha$. Specifically, label smoothing replaces the indicator value `1' (for the ground-truth class label) with $p = 1 -\alpha$, distributing the remaining $\alpha$ probability mass across all other class labels (Figure \ref{fig:soften_methods} left, ``Soft Target''). 
One can interpret label smoothing as accounting for the {\em average} loss of information resulting from averaging over transforms from the range $S$. From this perspective, the smoothing factor $\alpha$ can be written as a function of the {\em fixed} transform range $S$:
\begin{align}\label{eq:label_smoothing}
    (x_i,y_i) \Rightarrow \left(t_{\phi \sim S}(x_i), g_{\alpha(S)}(y_i)\right),\text{\bf Label Smoothing}
\end{align}

\textbf{Soft Augmentation}, our proposed approach, can now be described succinctly as follows: replace the fixed smoothing factor $\alpha(S)$ with an adaptive smoothing factor $\alpha(\phi)$, that depends on the degree of the {\em specific} sampled augmentation $\phi$ applied to $x_i$:
\begin{multline}
\label{eq:soft_augment}
       (x_i,y_i)\Rightarrow \left(t_{\phi \sim S}(x_i), g_{\alpha(\phi)}(y_i)\right), \\
       \text{\bf Soft Augmentation (Target)}
\end{multline}
Crucially, conditioning on the information loss from a particular $\phi$ allows one to define far {\em larger} augmentation ranges $S$. We will show that such a strategy consistently produces robust performance improvements with little tuning across a variety of datasets, models, and augmentation strategies. 

{\bf Extensions} to Soft Augmentation may be proposed by also considering loss reweighting~\cite{learning_reweight, yi2021reweighting}, which is an alternative approach for softening the impact of an augmented example by down-weighting its contribution to the loss. To formalize this, let us write the training samples of a supervised dataset as triples including a weight factor $w_i$ (that is typically initialized to all `1’s). One can then re-purpose our smoothing function $g$ to modify the weight instead of (or in addition to) the target label (Figure \ref{fig:soften_methods} left):
\begin{multline}
    \label{eq:soft_augment_w}
    (x_i,y_i,w_i) \Rightarrow  \left(t_{\phi \sim S}(x_i), y_i, g_{\alpha(\phi)}(w_i)\right), \\
    \text{\bf Soft Augmentation (Weight)}
\end{multline}
\begin{multline}
    \label{eq:soft_augment_tw}
    (x_i,y_i,w_i) \Rightarrow  \left(t_{\phi \sim S}(x_i), g_{\alpha(\phi)}(y_i),g_{\alpha(\phi)}(w_i)\right).\\
    \text{\bf Soft Augmentation (Target \& Weight)}
\end{multline}



Finally, one may wish to soften targets by exploiting class-specific confusions when applying $\alpha(\phi)$; 
the smoothed target label of a highly-occluded truck example could place more probability mass on other vehicle classes, as opposed to distributing the remaining probability equally across all other classes. Such extensions are discussed in Section \ref{sec:discussion}.

\section{Experiments} \label{sec:exp}

\subsection{Soft Augmentation with Crop}

\begin{figure*}[b]
    \centering

    \includegraphics[width=0.85\linewidth]{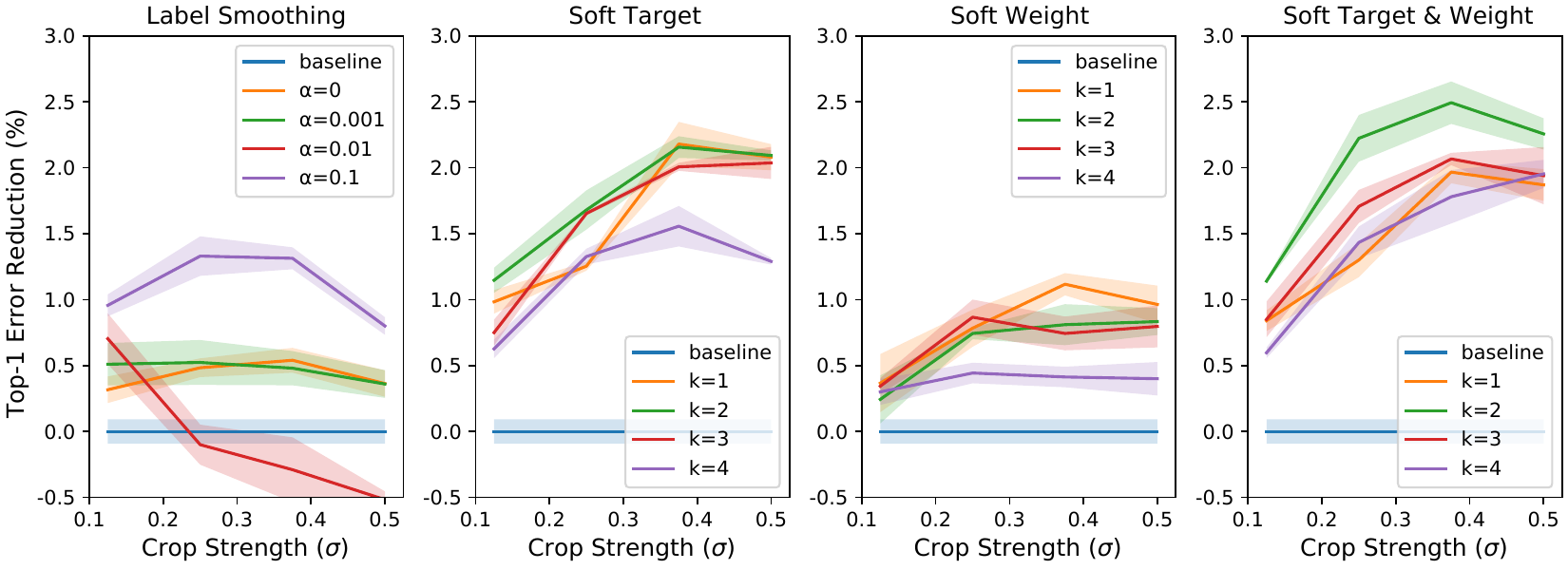}
    \caption{Soft Augmentation reduces the top-1 validation error of ResNet-18 on Cifar-100 by up to $2.5\%$ via combining both target and weight softening (Equation \ref{eq:soft_augment_tw}). Applying target softening alone (Equation~\ref{eq:soft_augment}) can boost performance by $\sim 2\%$. Crop parameters $tx,ty$ are independently drawn from $N(0,\sigma L)$ ($L=32$). Higher error reductions indicate better performance over baseline. All results are the means and standard errors across 3 independent runs. 
    }
    \label{fig:ablation}
    \vspace{-3mm}
\end{figure*}

As a concrete example of the proposed Soft Augmentation, we consider the crop transform $t_{(tx,ty,w,h)}(x)$. In the case of $32 \times 32$ pixel Cifar images~\cite{cifar}, the cropped images typically have a constant size $w = h = 32$, and $t(x)$ is fully parameterized by $tx$ and $ty$, which are translational offsets between the cropped and the original image. As shown in Figure \ref{fig:hard_vs_soft} (left), the standard hard crop augmentation for the Cifar-10/100 classification tasks draws $tx, ty$ independently from a uniform distribution of a modest range $U(-4,4)$. Under this distribution, the minimal visibility of an image is 77\% and ResNet-18 models trained on the Cifar-100 task achieve mean top-1 validation error of $20.80\%$ across three independent runs (Figure \ref{fig:hard_vs_soft} left). Naively applying aggressive hard crop augmentation drawn from a more aggressive range $U(-16,16)$ \emph{increases} top-1 error by $2.19\%$ (Figure \ref{fig:hard_vs_soft} middle). We make two changes to the standard crop augmentation.

We first propose drawing $tx,ty$ independently from a scaled normal distribution $S* \sim N(0,\sigma L)$ (with clipping such that $|tx|<L, |ty|<L$),  where $L$ is the length of the longer edge of the image ($L=32$ for Cifar). The distribution has zero mean and $\sigma$ controls the relative spread of the distribution hence the mean occlusion level. Following the $3\sigma$ rule of normal distribution, an intuitive tuning-free choice is to set $\sigma \approx 0.3$, where $ \sim 99\%$ of cropped samples have visibility $\geq 0$.  Figure \ref{fig:ablation} (left, $\alpha = 0$) shows that changing the distribution alone without target softening provides a moderate $\sim 0.4\%$ performance boost across crop strength $\sigma$.

Directly borrowing the findings from human vision reseach~\cite{tang2018recurrent}, one can define an adaptive softening $\alpha(tx,ty,k)$ that softens the ground truth learning target. Similar to Label Smoothing~\cite{label_smoothing}, a hard target can be softened to confidence $p \in [0,1]$. Instead of a fixed $\alpha$, consider a family of power functions that produces target hardness $p$ given crop parameters $tx, ty$ and curve shape $k$:
\begin{equation}
\label{eq:soften_curve}
    p = 1 - \alpha(tx,ty,k) = 1 - (1 - p_{min})(1 -v_{(tx,ty)})^{k},
\end{equation}
where $v_{(tx,ty)} \in [0,1]$ is the image visibility which is a function of $tx$ and $ty$. The power function family is a simple one-parameter formulation that allows us to test both linear ($k=1$) and non-linear ($k\neq 1$) softening: higher $k$ provides flatter plateaus in high visibility regime (see Figure \ref{fig:soften_methods} right). As seen in Label Smoothing, $p$ can be interpreted as ground truth class probability of the one-hot learning target. $p_{min}$ is the chance probability depending on the task prior. For example, for Cifar-100, $p_{min} = \frac{1}{\# classes} = 0.01$. 

Equation \ref{eq:soften_curve} has three assumptions: 1) the information loss is a function of image visibility and all information is lost only when the image is fully occluded, 2) the original label of a training image has a confidence of $100\%$, which suggests that there is no uncertainty in the class of the label, and 3) the information loss of all images can be approximated by a single confidence-visibility curve. While the first assumption is supported by observations of human visual classification research~\cite{tang2018recurrent}, and empirical evidence in Sections \ref{sec:ablation} and \ref{sec:supervised} suggests that the second and the third assumptions approximately hold, the limitations to these assumptions will be discussed in Section \ref{sec:discussion}.

\subsection{How to Soften Targets}\label{sec:ablation}

As prescribed by Equations \ref{eq:soft_augment}, \ref{eq:soft_augment_w}, and \ref{eq:soft_augment_tw}, three versions of Soft Augmentation are compared with Label Smoothing across a range of crop strength $\sigma$. The popular ResNet-18 models~\cite{resnetv2} are trained on the 100-class classification Cifar-100 training set. Top-1 error reductions on the validation set are reported (details in \ref{sec:appendix_exp}). Consistent with prior studies, label smoothing can boost model performance by $\sim 1.3 \%$ when the smoothing factor $\alpha$ is properly tuned (Figure \ref{fig:ablation} left).

Combining both target and weight softening (Equation \ref{eq:soft_augment_tw}) with $k=2$ and $\sigma \approx 0.3 $ boosts model performance by $ 2.5\%$  (Figure \ref{fig:ablation} right). Note that $k=2$ qualitatively resembles the shape of the curve of human visual confidence degradation under occlusion reported by Tang \etal~\cite{tang2018recurrent}. Interestingly, the optimal $\sigma \approx 0.3$ fits the intuitive $3$-$\sigma$ rule. In the next section we freeze $k=2$ and $\sigma = 0.3$ and show robust improvements that generalize to Cifar-10~\cite{cifar}, ImageNet-1K~\cite{imagenet}, and ImageNet-V2 tasks~\cite{imagenetv2}.

\subsection{Supervised Classification}\label{sec:supervised}

\subsubsection{Comparison with Related Methods}

\begin{table}[t]
    \centering
        \caption{Soft augmentation outperforms related methods. Optimal hyperparameters for Mixup~\cite{zhang2017mixup}, Cutout~\cite{cutout}, and Online Label Smoothing~\cite{ols} were applied. $\alpha$ of Focal Loss is tuned as \cite{lin2017focal} did not prescribe an optimal $\alpha$ for Cifar classification. It is worth noting that our baseline model (20.80\%) not only outperforms other published baseline models by 1.5\% to 4.8\%, but also beat various implementations of Mixup and Cutout. Top-1 errors of ResNet-18 on Cifar-100 are reported.}
    \resizebox{0.45\textwidth}{!}{
    \begin{tabular}{c c c}
        \toprule
         \multicolumn{2}{c}{ResNet-18} & {Top-1 Error} \\

         \midrule
         \multirow{4}{*}{Baseline} 
         & Zhang \etal~\cite{zhang2017mixup} & 25.6\\
         & DeVries and Taylor~\cite{cutout} & $22.46_{\pm 0.31}$ \\
         & Kim \etal~\cite{comixup} & 23.59\\
         & Ours & \bm{$20.80_{\pm 0.11}$} \\
         \midrule
         
         \multirow{3}{*}{Mixup} 
         & Zhang \etal~\cite{zhang2017mixup} & 21.1\\
         & Kim \etal~\cite{comixup} & 22.43\\
         & Ours & \bm{$19.88_{\pm 0.38}$} \\
         \midrule
         
         \multirow{2}{*}{Cutout} 
         & DeVries and Taylor~\cite{cutout} & $21.96_{\pm 0.24}$ \\
         & Ours & \bm{$20.51_{\pm 0.02}$} \\
         \midrule
         Label Smoothing & \multirow{7}{*}{Ours} & {$19.47_{\pm 0.18}$} \\
         Online Label Smoothing &   & $20.12_{\pm 0.05}$ \\
         Focal Loss ($\alpha = 1$) &   & $20.45_{\pm 0.08}$ \\
         Focal Loss ($\alpha = 2$)&   & $20.38_{\pm 0.08}$ \\
         Focal Loss ($\alpha = 5$) &   & $20.69_{\pm 0.17}$ \\
         RandAugment &   & $20.99_{\pm 0.11}$ \\
         Soft Augmentation &   & \bm{$18.31_{\pm 0.17}$} \\
        \bottomrule
    \end{tabular}
    }
    \vspace{-3mm}
    \label{tab:comparison}
\end{table}
As mentioned in Section \ref{sec:related}, many approaches similar to soft augmentation have demonstrated empirical performance gains, including additional data augmentation transforms~\cite{cutout}, learning augmentation policies~\cite{cubuk2020randaugment}, softening learning targets~\cite{label_smoothing}, and modifying loss functions~\cite{lin2017focal}. However, as training recipes continued to evolve over the past decade, baseline model performance has improved accordingly. As seen in Table \ref{tab:comparison} (Baseline), our baseline ResNet-18 models with a 500-epoch schedule and cosine learning rate decay~\cite{cosine} not only outperform many recent baseline models of the same architecture, but also beat various published results of Mixup and Cutout. To ensure fair comparisons, we reproduce 6 popular methods: Mixup, Cutout, Label Smoothing, Online Label Smoothing, Focal Loss, and RandAugment, and report the Top-1 Error on Cifar-100 in Table \ref{tab:comparison}. Additional comparisons with the self-reported results are available in Appendix Table \ref{tab:extended_comparison}. 

Table \ref{tab:comparison} shows that Soft Augmentation outperforms all other single methods. It is worth noting that although focal loss~\cite{lin2017focal} proposed for detection tasks, it can be tuned to slightly improve classification model performance. 

\subsubsection{Soft Augmentation Compliments RandAugment}
\begin{table*}[ht]
    \vspace{-5mm}
    \centering
        \caption{ Soft Augmentation (SA) with a \textbf{fixed} softening curve of $k=2$ \textbf{doubles} the top-1 error reduction of RandAugment (RA) across datasets and models. Note that the ResNet-18 models trained with SA on Cifar-10/100 even outperform larger baseline ResNet-50 and WideResNet-28 models. All results are mean $\pm$ standard error of top-1 validation error in percentage. Best results are shown in bold, runners-up are underlined, and results in parentheses indicate improvement over baseline. Statistics are computed from three runs.}
    
    \resizebox{0.95\textwidth}{!}{
    
    \begin{tabular}{c c c c c c}
        \toprule
         Dataset & Model & Baseline & SA & RA & SA+RA \\
         
        \midrule
         \multirow{5}{*}{Cifar100} & EfficientNet-B0 & $49.70_{\pm 1.55} $ & \underline{$42.13_{\pm 0.45} (-7.57)$} & $46.68_{\pm 1.52} (-3.02)$ & $\bm{38.72_{\pm 0.71} (-10.98)}$ \\
         
         & ResNet-18 & $20.80_{\pm 0.11} $ & \underline{$18.31_{\pm 0.17} (-2.49)$} & $20.99_{\pm 0.11} (+0.19)$ & $\bm{18.10_{\pm 0.20} (-2.70)}$ \\
         
          & ResNet-50 & $20.18_{\pm 0.30} $ & \underline{$18.06_{\pm 0.24} (-2.12)$} & $18.57_{\pm 0.09} (-1.61)$ & $\bm{16.72_{\pm 0.06} (-3.46)}$ \\
          
          & WideResNet-28 & $18.60_{\pm 0.19} $ & \underline{$16.47_{\pm 0.18} (-2.13)$} & $17.65_{\pm 0.14} (-0.95)$ & $\bm{15.37_{\pm 0.17} (-3.23)}$ \\
          
          & PyramidNet + ShakeDrop & $15.77_{\pm 0.17} $ & $14.03_{\pm 0.05} (-1.75)$ & \underline{$14.02_{\pm 0.28} (-1.76)$} & $\bm{12.78_{\pm 0.16} (-2.99)}$ \\
          
        \midrule  
         \multirow{5}{*}{Cifar10} & EfficientNet-B0 & $17.73_{\pm 0.69} $ & \underline{$12.21_{\pm 0.22} (-5.52)$} & $14.54_{\pm 0.47} (-3.19)$ & $\bm{11.67_{\pm 0.26} (-6.06)}$ \\
         
         & ResNet-18 & $4.38_{\pm 0.05} $ & \underline{$3.51_{\pm 0.08} (-0.87)$} & $3.89_{\pm 0.06} (-0.49)$ & $\bm{3.27_{\pm 0.08} (-1.11)}$ \\
         
          & ResNet-50 & $4.34_{\pm 0.14} $ & \underline{$3.67_{\pm 0.08} (-0.67)$} & $3.91_{\pm 0.14} (-0.43)$ & $\bm{3.01_{\pm 0.02} (-1.33)}$ \\
          
          & WideResNet-28 & $3.67_{\pm 0.08} $ & \underline{$2.85_{\pm 0.02} (-0.82)$} & $3.26_{\pm 0.04} (-0.41)$ & $\bm{2.45_{\pm 0.03} (-1.20)}$ \\
          
          & PyramidNet + ShakeDrop & $2.86_{\pm 0.03} $ & \underline{$2.26_{\pm 0.02} (-0.60)$} & $2.32_{\pm 0.08} (-0.54)$ & $\bm{2.02_{\pm 0.01} (-0.84)}$ \\

        \midrule  
         \multirow{2}{*}{ImageNet-1K} & ResNet-50 & $22.62_{\pm<0.01} $ & \underline{$21.66_{\pm0.02} (-0.96)$} & $22.02_{\pm0.02} (-0.60)$ & $\bm{21.27_{\pm0.05} (-1.35)}$ \\
         
         & ResNet-101 & $20.91_{\pm0.04} $ & $20.63_{\pm0.03} (-0.28)$ & \underline{$20.39_{\pm0.07} (-0.52)$} & $\bm{19.86_{\pm0.03} (-1.05)}$ \\
        \midrule
         \multirow{2}{*}{ImageNet-V2} & ResNet-50 & $34.97_{\pm0.03}$ & \underline{$ 33.32_{\pm0.10} (-1.65)$} & $34.16_{\pm0.21}  (-0.81)$ & $\bm{32.38_{\pm0.16} (-2.59)}$ \\
         & ResNet-101 & $32.68_{\pm0.04} $ & \underline{$31.81_{\pm0.16} (-0.87)$} & $32.08_{\pm0.19} (-0.60)$ & $\bm{31.26_{\pm0.12} (-1.42)}$ \\
        \bottomrule
    \end{tabular}
    }
    \label{tab:supervised}
    \vspace{-2mm}
\end{table*}

This section investigates the robustness of Soft Augmentation across models and tasks, and how well it compares or complements more sophisticated augmentation policies such as RandAugment~\cite{cubuk2020randaugment}. The ImageNet-1K dataset~\cite{imagenet} has larger and variable-sized images compared to the Cifar~\cite{cifar} datasets. In contrast with the fixed-sized crop augmentation for Cifar, a crop-and-resize augmentation $t_{(tx,ty,w,h)}(x)$ with random location $tx,ty$ and random size $w,h$ is standard for ImageNet training recipes~\cite{googlenet,cubuk2019autoaugment,cubuk2020randaugment}. The resizing step is necessary to produce fixed-sized training images (e.g. $224 \times 224$). We follow the same $\sigma = 0.3 $ principle for drawing $tx,ty$ and $w,h$ (details in \ref{sec:appendix_exp}).

Comparing single methods, Soft Augmentation with \emph{crop only} consistently outperforms the more sophisticated RandAugment with 14 transforms (Table \ref{tab:supervised}). The small ResNet-18 models trained with SA on Cifar-10/100 even outperforms much larger baseline ResNet-50~\cite{imagenetv2} and WideResNet-28~\cite{zagoruyko2016wide} models. 

Because RandAugment is a searched policy that is originally prescribed to be applied in addition to the standard crop augmentation~\cite{cubuk2020randaugment}, one can easily replace the standard crop with soft crop and combine Soft Augmentation and RandAugment. As shown in Table \ref{tab:supervised}, Soft Augmentation complements RandAugment by doubling its top-1 error reduction across tasks and models. 

Note that for the small ResNet-18 model trained on Cifar-100, the fixed RandAugment method slightly degrades its performance. Consistent with observations from Cubuk \etal~\cite{cubuk2020randaugment}, the optimal hyperparameters for RandAugment depend on the combination of model capacity and task complexity. Despite the loss of performance of applying RandAugment alone, adding Soft Augmentation reverses the effect and boosts performance by $2.7 \%$. 

For the preceding experiments, a fixed $k=2$ is used for Soft Augmentation and the official PyTorch RandomAugment~\cite{pytorch} is implemented to ensure a fair comparison and to evaluate robustness. It is possible to fine-tune the hyperparameters for each model and task to achieve better empirical performance.  
\vspace{-3mm}

\subsubsection{Occlusion Robustness}

\begin{figure*}[h]
    \vspace{-4mm}
    \centering
    \includegraphics[width=0.99\linewidth]{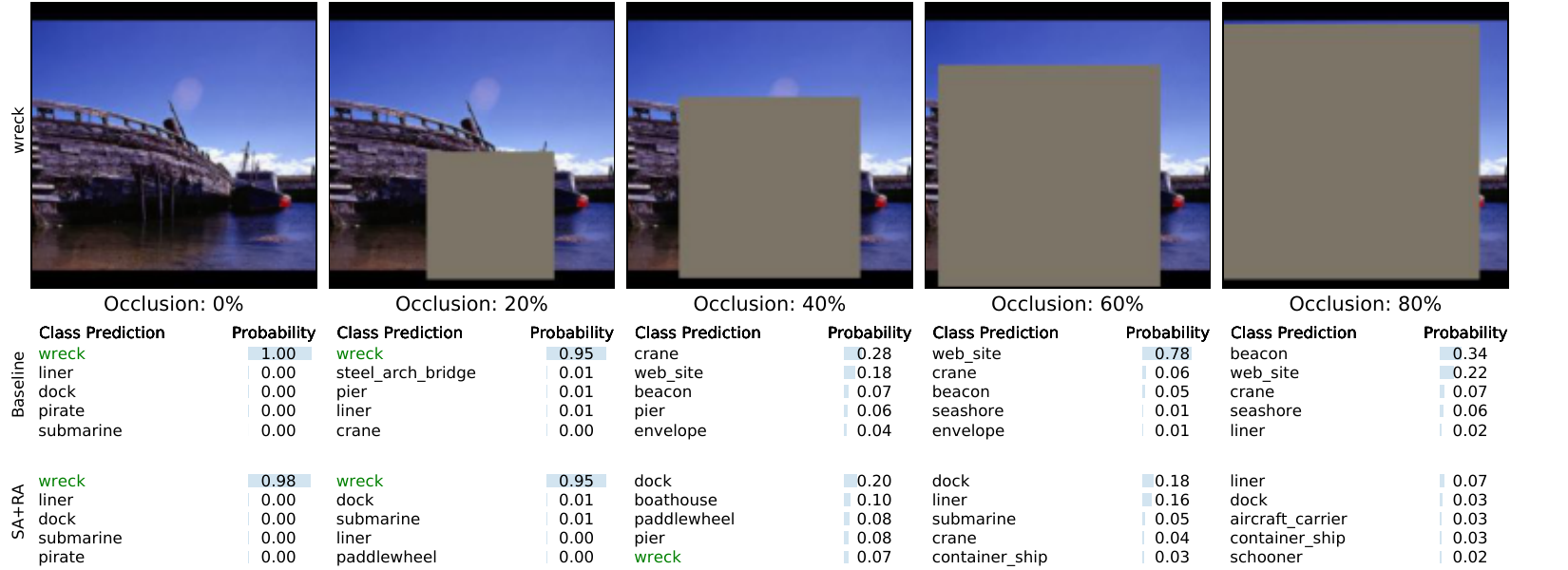}
    \caption{Examples of occluded ImageNet validation images and model predictions of ResNet-50. $224 \times 224$ validation images of ImageNet are occluded with randomly placed square patches that cover $\lambda$ of the image area. $\lambda$ is set to $\{0\%,20\%,40\%,60\%,80\% \}$ to create a range of occlusion levels.}
    \label{fig:occlusion_example}
    \vspace{-2mm}
\end{figure*}

As discussed in Section \ref{sec:related}, occlusion robustness in both human vision~\cite{tang2018recurrent,zhu2019robustness, lowe1999invariant_features} and computer vision~\cite{wang2020robust,wyatte2012limits,kortylewski2020compositional} have been an important property for real world applications of vision models as objects. To assess the effect of soft augmentation on occlusion robustness of computer vision models, ResNet-50 models are tested with occluded ImageNet validation images (Figure \ref{fig:occlusion_example} and Appendix Figure \ref{fig:occlusion_example_additional}). $224 \times 224$ validation images of ImageNet are occluded with randomly placed square patches that cover $\lambda$ of the image area. $\lambda$ is set to $\{0\%,20\%,40\%,60\%,80\% \}$ to create a range of occlusion levels. 

\begin{figure}[h]
    \vspace{-4mm}
    \centering
    \includegraphics[width=1\linewidth]{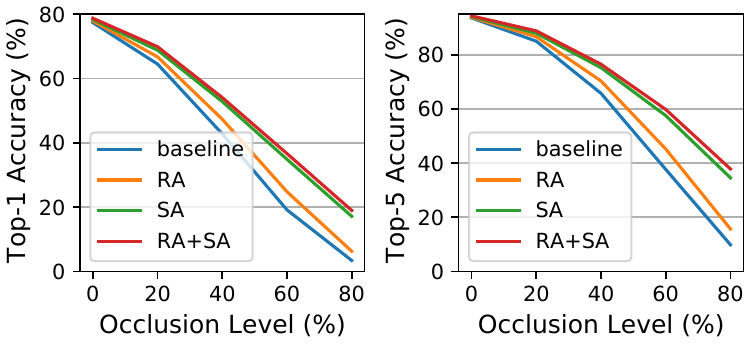}
    \caption{Soft Augmentation improves occlusion robustness of ResNet-50 on ImageNet. Both RandAugment (RA) and Soft Augmentation (SA) improve occlusion robustness independently. Combining RA with SA reduces Top-1 error by up to 17\%. At 80\% occlusion level, compared with baseline accuracy (3.42\%), SA+RA achieves more than $\bm{4\times}$ \textbf{accuracy} (18.98\%).}
    \label{fig:occlusion_curves}
    \vspace{-3mm}
\end{figure}

As shown in Figure \ref{fig:occlusion_curves}, both RandAugment (RA) and Soft Augmentation (SA) improve occlusion robustness independently across occlusion levels. Combining RA with SA reduces Top-1 error by up to 17\%. At 80\% occlusion level, SA+RA achieves more than $\bm{4\times}$ \textbf{accuracy improvement} over the baseline (18.98\% vs 3.42\%).
\vspace{-3mm}

\subsubsection{Confidence Calibration}

In addition to top-1 errors, reliability is yet another important aspect of model performance. It measures how close a model's predicted probability (confidence) tracks the true correctness likelihood (accuracy).  Expected Calibration Error (ECE) is a popular metric~\cite{guo2017calibration, ensemble, ddu} to measure confidence calibration by dividing model predictions into $M$ confidence bins ($B_m$) and compute a weighted average error between accuracy and confidence:
\vspace{-3mm}
\begin{equation}
    ECE = \sum_{m=1}^{M}{\frac{|B_m|}{n}|acc(B_m) - conf(B_m)|},
    \vspace{-3mm}
\end{equation}
where $n$ is the number of samples, $acc(B_m)$ denotes the accuracy of bin $m$, and $conf(B_m)$ denotes mean model confidence of bin $m$. Consistent with Guo \etal~\cite{guo2017calibration}, we set $M=10$ and compute ECE for Cifar-10 and Cifar-100 tasks.

As shown in Table \ref{tab:calibration}, many methods \cite{ddu, sngp, duq, ensemble} have been proposed to improve confidence calibration, sometimes at the cost of drastically increased computational overhead~\cite{ensemble}, or degraded raw performance~\cite{sngp}. We show in Table \ref{tab:calibration} (and Appendix Table \ref{tab:ece_imagenet}) that it is possible to further reduce model top-1 error and expected calibration error simultaneously. 

Compared to previous single-model methods, our strong baseline WideResNet-28 models achieves lower top-1 error at the cost of higher ECE. Combining Soft Augmentation with more recently developed augmentation policies such as TrivialAugment~\cite{muller2021trivialaugment} (SA+TA) reduces top-1 error by $4.36 \%$ and reduces ECE by more than half on Cifar-100, outperforming the $4 \times$ more computationally expensive 5-ensemble model~\cite{ensemble}. To the best of our knowledge, this is state of the art ECE performance for WideResNet-28 on Cifar without post-hoc calibration. 

\begin{table}[h]
    \vspace{-3mm}
    \centering
    
        \caption{Soft Augmentation improves both accuracy and calibration. We report mean and standard error of three WideResNet-28 runs per configuration (bottom two rows). 
        On the more challenging Cifar-100 benchmark, our Baseline already outperforms much of prior work in terms of Top-1 error, but has worse calibration error (ECE). 
        Applying Soft Augment + Trivial Augment (SA+TA) reduces Top-1 error by 4.36\% and reduces ECE by \textbf{more than half}, outperforming even compute-heavy models such as the 5-Ensemble ~\cite{ensemble}. Similar trends hold for Cifar-10.}
        \vspace{-3mm}
    \resizebox{0.48\textwidth}{!}{
    
    \begin{tabular}{c cc cc}
        \toprule
         Method & \multicolumn{2}{c}{Cifar-100} & \multicolumn{2}{c}{Cifar-10} \\
        \midrule
          & \underline{Top-1 Error} & \underline{ECE} & \underline{Top-1 Error} & \underline{ECE} \\

         Energy-based~\cite{liu2020energy} & 19.74 & 4.62 & 4.02 & 0.85\\
         
         DUQ~\cite{duq} & -- & -- & 5.40 & 1.55 \\
         
         SNGP~\cite{sngp} & 20.00 & 4.33 & 3.96 & 1.80\\
         
         DDU~\cite{ddu} & 19.02 & 4.10 & 4.03 & 0.85\\
         
         5-Ensemble~\cite{ensemble} & \underline{17.21} & \underline{3.32} & \underline{3.41} & \underline{0.76}\\
         
        \midrule
         Our Baseline & $18.60_{\pm 0.16}$ & $4.86_{\pm 0.10}$ & $3.67_{\pm 0.07}$ & $2.22_{\pm 0.03}$\\
         SA+TA & $\bm{14.24_{\pm 0.11}}$ & $\bm{1.76_{\pm 0.15}}$ & $\bm{2.23_{\pm 0.06}}$ & $\bm{0.61_{\pm 0.10}}$ \\
        \bottomrule
    \end{tabular}
    }
    \label{tab:calibration}
    \vspace{-2mm}
\end{table}

\subsection{Soft Augmentation Boosts Self-Supervised Learning}

In contrast with supervised classification tasks where the learning target $y_i$ is usually a one-hot vector, many self-supervised methods such as SimSiam~\cite{simsiam} and Barlow Twins~\cite{barlow}  learn visual feature representations without class labels by encouraging augmentation invariant feature representations. This section investigates whether Soft Augmentation generalizes to learning settings where no one-hot style labeling is provided. 

In a typical setting, two random crops of the same image are fed into a pair of identical twin networks (e.g., ResNet-18) with shared weights and architecture. The learning target can be the maximization of similarity between the feature representations of the two crops~\cite{simsiam}, or minimization of redundancy~\cite{barlow}. By default, all randomly cropped pairs have equal weights. We propose and test two alternative hypotheses for weight softening with SimSiam. To accommodate self-supervised learning,  Equation \ref{eq:soften_curve} is modified by replacing visibility $v_{tx,ty}$ with intersection over union $IoU$ of two crops of an image:
\vspace{-3mm}
\begin{multline}\label{eq:ssl_hypothesis_1}
    \vspace{-0mm}
    p = 1 - \alpha(\phi_1, \phi_2,k) = 1 - (1 - p_{min})(1 - IoU_{\phi_1, \phi_2})^{k}, \\
    \text{\bf SA\#1}
\end{multline}
where $\phi_1 = (tx_1,ty_1,w_1,h_1)$ and $\phi_2 = (tx_2,ty_2,w_2,h_2)$ are crop parameters for the first and second sample in a pair. $p$ is used to soften weights only as no one-hot classification vector is available in this learning setting.
With this hypothesis (SA\#1), ``hard" sample pairs with low IoUs are assigned low weights. Alternatively, one can assign lower weights to ``easy" sample pairs with higher IoUs (SA\#2), as prescribed by Equation \ref{eq:ssl_hypothesis_2}:
\vspace{-3mm}
\begin{multline}
    \label{eq:ssl_hypothesis_2}
    p = 1 - \alpha(\phi_1, \phi_2, k) = 1 - (1 - p_{min})(IoU_{\phi_1, \phi_2})^{k}.\\
    \text{\bf SA\#2}
\end{multline}
We first test all three hypotheses (baseline, SA\#1, and SA\#2) on Cifar-100 with the SimSiam-ResNet-18 models. Table \ref{tab:ssl} (top) shows that SA\#1 outperform both baseline and SA\#2 (details in \ref{sec:ssl_appendix}). Additional experiments show that models trained with the same SA\#1 configuration also generalize to Cifar-10 (Table \ref{tab:ssl} bottom).
\begin{table}[t]
    \centering
        \caption{Soft Augmentation (SA\#1) improves self supervised learning with SimSiam (ResNet-18) on Cifar-100 by down-weighting sample pairs with \emph{small} intersection over union (IoU), outperforming the opposite hypothesis (SA\#2) of down-weighting pairs with \emph{large} IoU. For each configuration, we report means and standard errors of 3 runs with best learning rates (LR) found for Cifar-100. The effect of SA\#1 (with a fixed $k=4$) generalizes to Cifar-10 without re-tuning. 
        }
    \resizebox{0.48\textwidth}{!}{
    
    \begin{tabular}{c c c c c c c}
        \toprule
         Task &  LR & Baseline & LR & SA\#1 & LR & SA\#2 \\
        \midrule
         \multirow{1}{*}{Cifar100} & 0.2 & $37.64_{\pm 0.06}$ & 0.2 & $\bm{36.61}_{\pm 0.05}$ & 0.1 & $37.39_{\pm 0.06}$\\
          
        \midrule  
         \multirow{1}{*}{Cifar10} & 0.2 & $9.87_{\pm 0.03}$ & 0.2 & $\bm{9.31_{\pm 0.01}}$  & - & -\\
        \bottomrule
    \end{tabular}
    }
    \label{tab:ssl}
    \vspace{-0mm}
\end{table}

\section{Discussion} \label{sec:discussion}


{\bf Other augmentations.} While we focus on crop augmentations as an illustrative example, Soft Augmentation can be easily extended to a larger repertoire of transforms such as affine transforms and photometric distortions, as seen in the more sophisticated augmentation policies such as RandAugment. As the formulation of Equation \ref{eq:soften_curve} (and Figure \ref{fig:soften_methods} right) is directly inspired by the qualitative shape of human vision experiments from Tang \etal~\cite{tang2018recurrent}, optimal softening curves for other transforms may be discovered by similar human experiments. However, results with a second transform in Appendix Table \ref{tab:noise} suggest that Equation \ref{eq:soften_curve} generalizes to additive noise augmentation as well. A potential challenge is determining the optimal softening strategy when a combination of several transforms are applied to an image since the cost of a naive grid search increases exponentially with the number of hyperparameters. Perhaps reinforcement learning methods as seen in RandAugment can be used to speed up the search. 


{\bf Other tasks.} While we limit the scope of Soft Augmentation to image classification as it is directly inspired by human visual classification research, the idea can be generalized to other types of tasks such as natural language modeling and object detection. Recent studies have shown that detection models benefit from  soft learning targets in the final stages \cite{bodla2017soft_nms, li2020gfl}, Soft Augment has the potential to complement these methods by modeling information loss of image transform in the models' input stage.

{\bf Class-dependant augmentations.} As pointed out by Balestriero \etal~\cite{balestriero2022effects}, the effects of data augmentation are class-dependent. Thus assumption 3 of Equation \ref{eq:soften_curve} does not exactly hold. One can loosen it by adaptively determining the range of transform and softening curve on a per class or per sample basis. As shown in Equation \ref{eq:adaptive_soft_augment}, 
\begin{equation}\label{eq:adaptive_soft_augment}
(x_i,y_i) \Rightarrow \left(t_{\phi \sim S(x_i,y_i)}(x_i), g_{\alpha(\phi,x_i,y_i)}(y_i)\right),
\end{equation}
two adaptive improvements can be made: 1) the transformation range  $S$ where $\phi$ is drawn from can be made a function of sample $(x_i,y_i)$, 2) the softening factor $\alpha$ can also adapt to $(x_i,y_i)$. Intuitively, the formulation recognizes the heterogeneity of training samples of images at two levels. Firstly, the object of interest can occupy different proportions of an image. For instance, a high-resolution training image with a small object located at the center can allow more aggressive crop transforms without losing its class invariance. Secondly, texture and shape may contribute differently depending on the visual class. A heavily occluded tiger may be recognized solely by its distinctive stripes; in contrast, a minimally visible cloak can be mistaken as almost any clothing.

\section{Conclusion}

In summary, we draw inspiration from human vision research, specifically how human visual classification performance degrades non-linearly as a function of image occlusion. We propose generalizing data augmentation with invariant transforms to Soft Augmentation where the learning target (e.g. one-hot vector and/or sample weight) softens non-linearly as a function of the degree of the transform applied to the sample. 

Using cropping transformations as an example, we empirically show that Soft Augmentation offers robust top-1 error reduction across Cifar-10, Cifar-100, ImageNet-1K, and ImageNet-V2. With a fixed softening curve, Soft Augmentation doubles the top-1 accuracy boost of the popular RandAugment method across models and datasets, and improves performance under occlusion by up to $4\times$. Combining Soft Augment with the more recently developed TrivialAugment further improves model accuracy and calibration simultaneously, outperforming even compute-heavy 5-ensemble models. Finally, self-supervised learning experiments demonstrate that Soft Augmentation also generalizes beyond the popular supervised one-hot classification setting.

{\clearpage
\small
\bibliographystyle{ieee_fullname}
\bibliography{main.bib}
}

\clearpage
\appendix
\renewcommand{\thesection}{Appendix \Alph{section}}

\section{Implementation}\label{sec:implementation}
\begin{lstlisting}[language=Python, caption=Pytorch implementation of Soft Crop Augmentation for Cifar.]
import torch

class SoftCropAugmentation: 
    def __init__(self, n_class, sigma=0.3, k=2):
        self.chance = 1/n_class
        self.sigma = sigma
        self.k = k

    def draw_offset(self, limit, sigma=0.3, n=100):
        # draw an integer from a (clipped) Gaussian
        for d in range(n):
            x = torch.randn((1))*sigma
            if abs(x) <= limit:
                return int(x)
        return int(0)

    def __call__(self, image, label):
        # typically, dim1 = dim2 = 32 for Cifar
        dim1, dim2 = image.size(1), image.size(2)
        # pad image
        image_padded = torch.zeros((3, dim1 * 3, dim2 * 3)) 
        image_padded[:, dim1:2*dim1, dim2:2*dim2] = image 
        # draw tx, ty
        tx = self.draw_offset(dim1, self.sigma_crop * dim1)
        ty = self.draw_offset(dim2, self.sigma_crop * dim2)
        # crop image
        left, right = tx + dim1, tx + dim1 * 2
        top, bottom = ty + dim2, ty + dim2 * 2
        new_image = image_padded[:, left: right, top: bottom]
        # compute transformed image visibility and confidence
        v = (dim1 - abs(tx)) * (dim2 - abs(ty)) / (dim1 * dim2)
        confidence = 1 - (1 - self.chance) * (1 - v) ** self.k
        return new_image, label, confidence
\end{lstlisting}

\begin{lstlisting}[language=Python, caption=Pytorch implementation of Soft Target loss function.]
import torch
import torch.nn.functional as F

def soft_target(pred, label, confidence):
    log_prob = F.log_softmax(pred, dim=1)
    n_class = pred.size(1)
    # make soft one-hot target
    one_hot = torch.ones_like(pred) * (1 - confidence) / (n_class - 1)
    one_hot.scatter_(dim=1, index=label, src=confidence)
    # compute weighted KL loss
    kl = confidence * F.kl_div(input=log_prob, 
                               target=one_hot, 
                               reduction='none').sum(-1)
    return kl.mean()
\end{lstlisting}






\section{Experiment Details}\label{sec:appendix_exp}

\subsection{Supervised Cifar-10/100}

For Cifar-100 experiments, we train all ResNet-like models with a batch size 128 on a single Nvidia V100 16GB GPU on Amazon Web Services (AWS) and with an intial learning rate 0.1 with cosine learning rate decay over 500 epochs. EfficientNet-B0 is trained with an initial learning rate of 0.025, PyramidNet-272 is trained with 2 GPUs. We use the Conv-BatchNorm-ReLU configuration of ResNet models~\cite{resnetv2} and WideResNet-28 with a widening factor of 10~\cite{zagoruyko2016wide}. Horizontal flip is used in all experiments as it is considered a lossless transformation in the context of Cifar images. We find decaying crop aggressiveness towards the end of training (e.g., last 20 epochs) by a large factor (e.g., reducing $\sigma$ by $1000 \times$) marginally improve performance on Cifar-100, but slightly hurts performance on Cifar-10. Accordingly, we only apply $\sigma$ decay for all Cifar-100 experiments. A single run of ResNet-18, ResNet-50, and WideResNet-28 takes $\sim 2.5$, $\sim 7$, $\sim 9$ GPU hours on Cifar-10/100, respectively.

\begin{figure}[h!]
    \centering

    \includegraphics[width=1.0\linewidth]{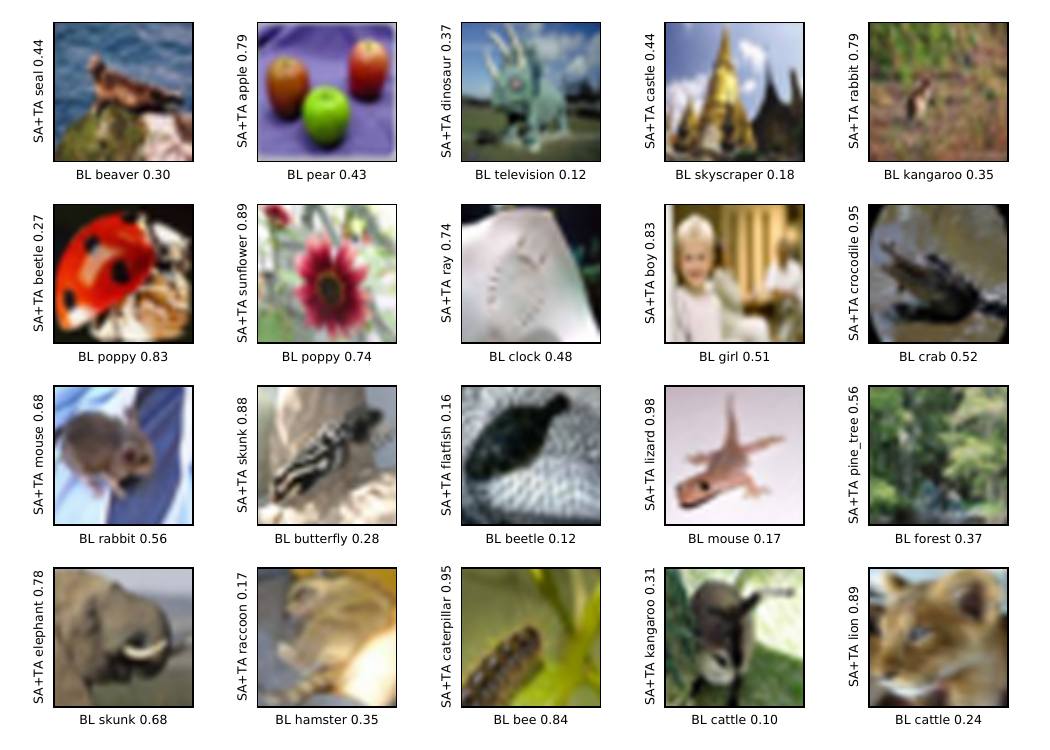}
    \vspace{-6mm}
    \caption{Example images of the Cifar-100 validation set and predictions of WideResNet-28. Predicted classes and confidence levels of models trained with Soft Augmentation + Trivial Augment (SA+TA) and baseline (BL) augmentation are reported. In many cases, SA+TA not only corrects the class prediction, but also improves the model confidence. For instance, BL mistakes ``seal'' for ``beaver'' (top-left, both classes belong to the same ``aquatic mammal'' superclass), and SA+TA makes a correct class prediction with higher confidence. 
    }
    \label{fig:examples}
    \vspace{-4mm}
\end{figure}
\clearpage

\subsection{Additional Results}

\begin{table}[h]
    \vspace{-5mm}
    \centering
        \caption{Comparing SA with other methods. Recommended hyperparameters for Mixup~\cite{zhang2017mixup}, Cutout~\cite{devries2017improved}, and Online Label Smoothing~\cite{ols}. $\alpha$ of Focal Loss is tuned as Lin \etal~\cite{lin2017focal} did not prescribe an optimal $\alpha$ for Cifar classification. Top-1 errors of ResNet-18 on Cifar-100 are reported.}
    \resizebox{0.3\textwidth}{!}{
    \begin{tabular}{c c}
        \toprule
         ResNet-18 &  Top-1 Error \\
         \midrule
         \multicolumn{1}{c}{\underline{ Zhang \etal~\cite{zhang2017mixup} } } & \\
         Baseline & 25.6 \\
         Mixup & 21.1 \\
         \midrule
         \multicolumn{1}{c}{Kim \etal~\underline{~\cite{puzzlemix} } } & \\
         Baseline & 23.67 \\
         Mixup & 23.16 \\
         Manifold Mixup & 20.98 \\
         Puzzle Mix & 19.62 \\
         \midrule
         \multicolumn{1}{c}{Kim \etal~\underline{~\cite{comixup} } } & \\
         Baseline & 23.59 \\
         Mixup & 22.43 \\
         Manifold Mixup & 21.64 \\
         Puzzle Mix & 20.62 \\
         Co-Mixup & 19.87 \\
         \midrule
         Our Baseline & $20.80_{\pm 0.11}$ \\
         Label Smoothing & $19.47_{\pm 0.18}$ \\
         Online Label Smoothing & $20.12_{\pm 0.05}$ \\
         Focal Loss ($\alpha = 1$) & $20.45_{\pm 0.08}$ \\
         Focal Loss ($\alpha = 2$)& $20.38_{\pm 0.08}$ \\
         Focal Loss ($\alpha = 5$) & $20.69_{\pm 0.17}$ \\
         Mixup ($\alpha = 1.0$) & $19.88_{\pm 0.38}$ \\
         Cutout ($L=8$) & $20.51_{\pm 0.02}$ \\
         SA & $18.31_{\pm 0.17}$ \\
         RA & $20.99_{\pm 0.11}$ \\
         SA + RA & $\bm{18.10_{\pm 0.20}}$ \\
        \bottomrule
    \end{tabular}
    }
    \vspace{-6mm}
    \label{tab:extended_comparison}
\end{table}

\begin{table}[h]
    \centering
        \caption{Soft Augmentation with additive noise improves ResNet-18 performance on Cifar-100. Given an image $X$ and a random noise pattern $X_{noise}$, and augmented image is given by $X_{aug}=X + \alpha X_{noise}$, where $\alpha$ is drawn from $N(0,0.1)$ and pixel values of $X_{noise}$ are also independently drawn from $N(0,0.1)$. Applying Soft Augmentation to additive noise boost performance over baseline as well as Soft Augmentation Crop + RandAugment.}
    \resizebox{0.45\textwidth}{!}{
    \begin{tabular}{c c}
        \toprule
         ResNet-18 &  Top-1 Error \\
         \midrule
         Baseline  & $ 20.80{\pm 0.11} $ \\
         RA & $20.99_{\pm 0.11}$ \\
         Hard Crop & $20.26_{\pm 0.12} $ \\
         SA-Crop (k=2) & $18.31_{\pm 0.17} $ \\
         Hard Noise & $20.68_{\pm 0.05} $ \\
         SA-Noise (k=1)  & $ 19.20{\pm 0.20} $ \\
         SA-Crop (k=2) + RA & $ 18.10_{\pm 0.20} $ \\
         SA-Noise (k=1) + SA-Crop (k=2) + RA & $ \bm{17.87{\pm 0.17}} $ \\
        \bottomrule
    \end{tabular}
    }
    \vspace{-4mm}
    \label{tab:noise}
\end{table}

\begin{table}[h]
    \vspace{-3mm}
    \centering
    
        \caption{Soft Augmentation reduces expected calibration error (ECE) of ResNet-50 on ImageNet. }
        \vspace{-3mm}
    \resizebox{0.4\textwidth}{!}{
    \begin{tabular}{c c c c c}
        \toprule
         Dataset &  Baseline & RA & SA & RA+SA \\ 
        \midrule
         ImageNet-1K & 5.11 & 4.09 & 3.17 & 2.78 \\

         ImageNet-V2 & 9.91 & 8.84 & 3.24 & 3.18 \\
        \bottomrule
    \end{tabular}
    }
    \label{tab:ece_imagenet}
    \vspace{-4mm}
\end{table}


\subsection{ImageNet}

All ImageNet-1k experiments are conducted with a batch size of 256 distributed across 4 Nvidia V100 16GB GPUs on AWS. The ImageNet Large Scale Visual Recognition Challenge (ILSVRC) 2012 dataset (BSD 3-Clause License) is downloaded from the official website (https://www.image-net.org/). Horizontal flip is used in all experiments as an additional lossless base augmentation. The base learning rate is set to 0.1 with a 5-epoch linear warmup and cosine decay over 270 epochs. A single run of ResNet-50 training takes $\sim 4 \times 4 = 16$ GPU days and ImageNet experiments take a total of $~600$ GPU days.

We use the official PyTorch~\cite{pytorch} implementation of RandAugment and ResNet-50/101 (BSD-style license) and run all experiments with the standard square input $L_{input} = W = H = 224$. Note that the original RandAugment~\cite{cubuk2020randaugment} uses a larger input size of $H=224,W=244$, but our re-implemention improved top-1 error (22.02 vs 22.4) of ResNet-50 despite using a smaller input size. ImageNet-V2 is a validation set proposed by He \etal~\cite{imagenetv2}.

For training, the standard crop transform has 4 hyperparameters: $(scale_{min}, scale_{max})$ define the range of the relative size of a cropped image to the original one, $(ratio_{min}, ratio_{max})$ determine lower and upper bound of the aspect ratio of the cropped patch before the final resize step. In practice, a scale is drawn from a uniform distribution $U(scale_{min}, scale_{max})$, then the logarithm of the aspect ratio is drawn from a uniform distribution $U(log(ratio_{min}),log(ratio_{max}))$. Default values are $scale_{min} =0.08, scale_{max} = 1.0, ratio_{min} = 3/4, ratio_{max} = 4/3 .$

Similar to our Cifar crop augmentation, we propose a simplified ImageNet crop augmentation with only 2 hyperparameters $\sigma, L_{min}$. First, we draw $\Delta w, \Delta h$ from a clipped rectified normal distribution $N^R(0, \sigma (L - L_{min}) )$ and get $w = W - \Delta w, h = H - \Delta h$
$L_{min}$ is the minimum resolution of a cropped image and set to half of input resolution $224$. $tx,ty$ are then independently drawn from $N(0, \sigma (W +w )), N(0, \sigma (H + h )).$ Note that we use a fixed set of intuitive values $\sigma=0.3, L_{min}= 1/2 L_{input} = 112$ for all the experiments.

For model validation, standard augmentation practice first resizes an image so that its short edge has length $L_{input} = 256$, then a center $224 \times 224$ crop is applied. Note that $L_{input}$ is an additional hyperparameter introduced by the test augmentation. In contrast, we simplify this by setting $L_{input}$ to the final input size $224$ and use this configuration for all ImageNet model evaluation.
\clearpage
\begin{figure*}[t]
    \centering
    \includegraphics[width=0.7\linewidth]{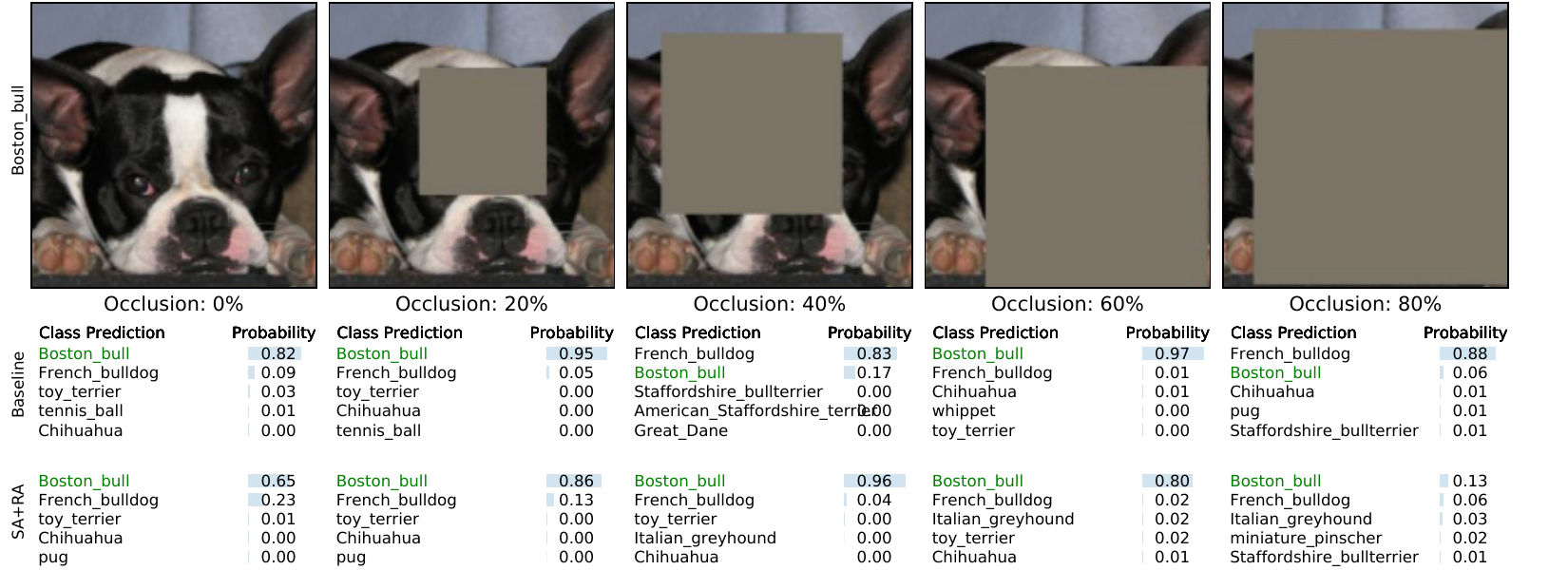}
    \includegraphics[width=0.7\linewidth]{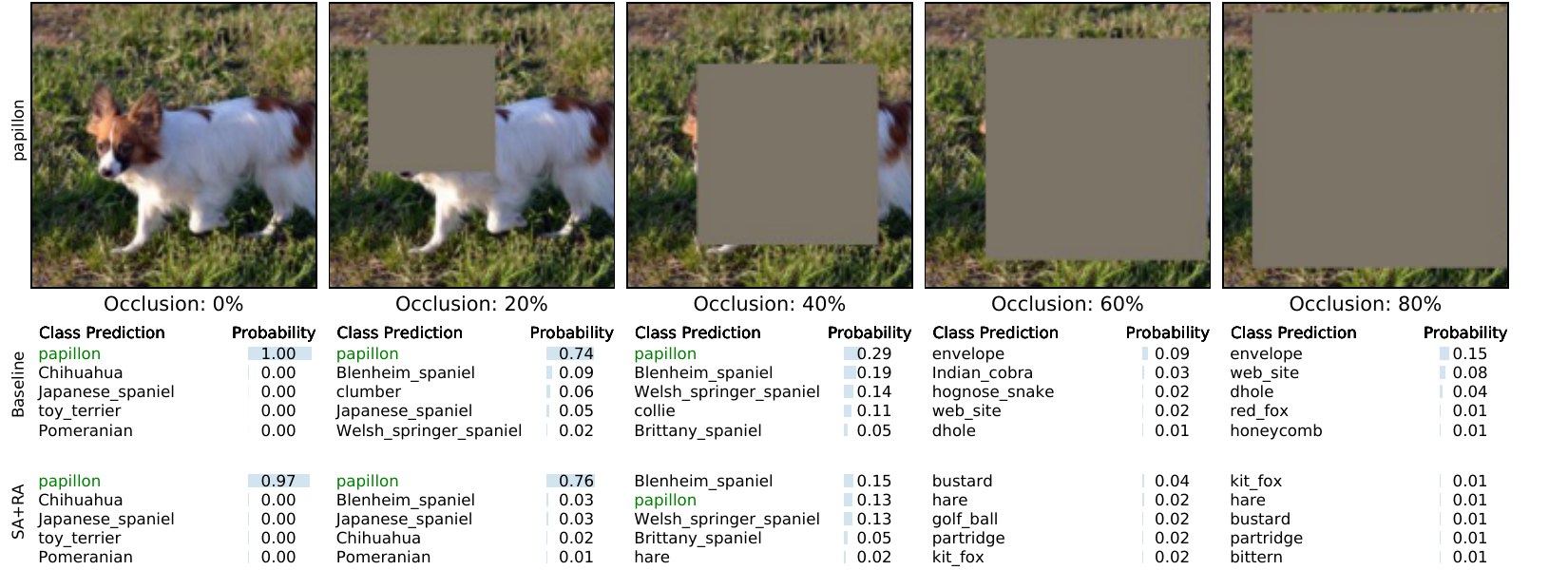}
    \includegraphics[width=0.7\linewidth]{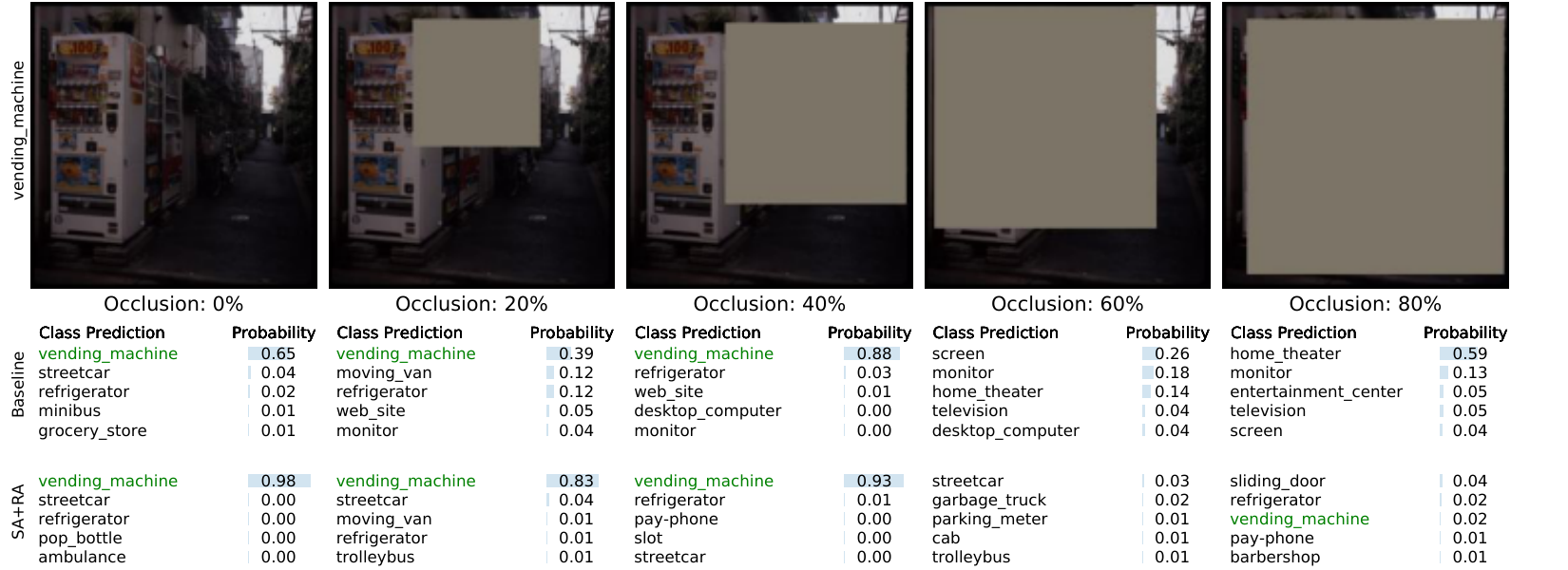}
    \includegraphics[width=0.7\linewidth]{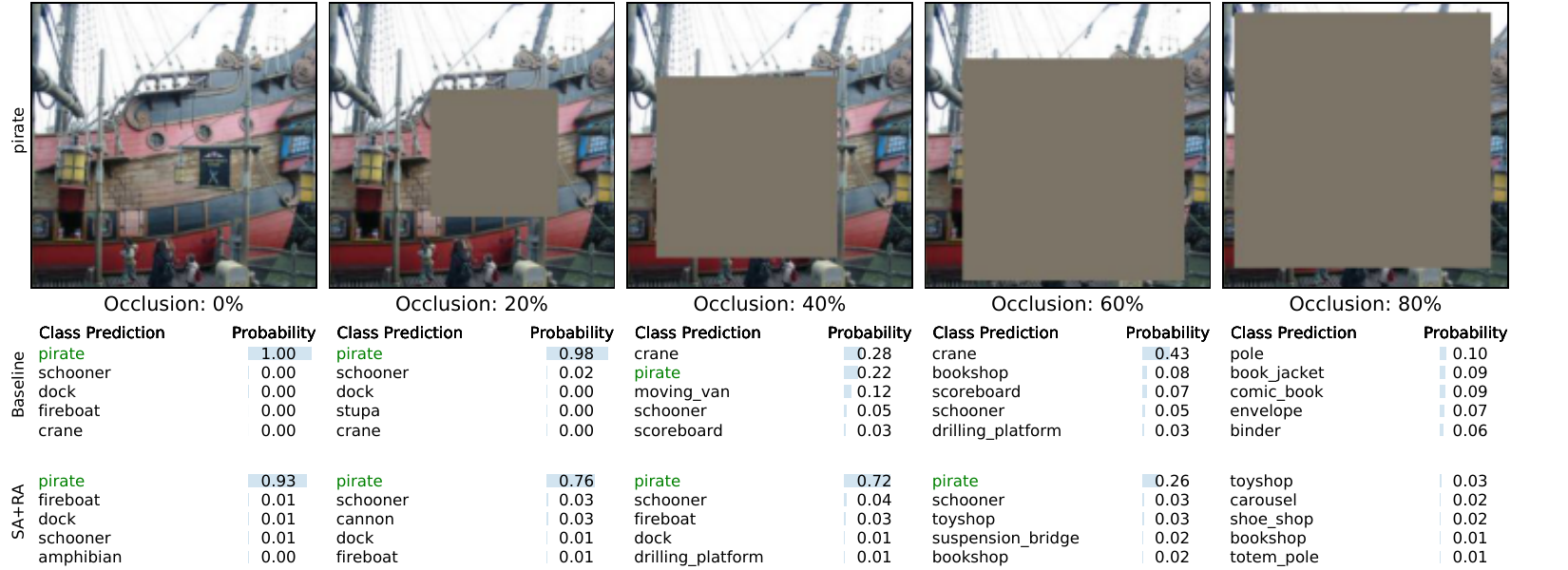}
    \includegraphics[width=0.7\linewidth]{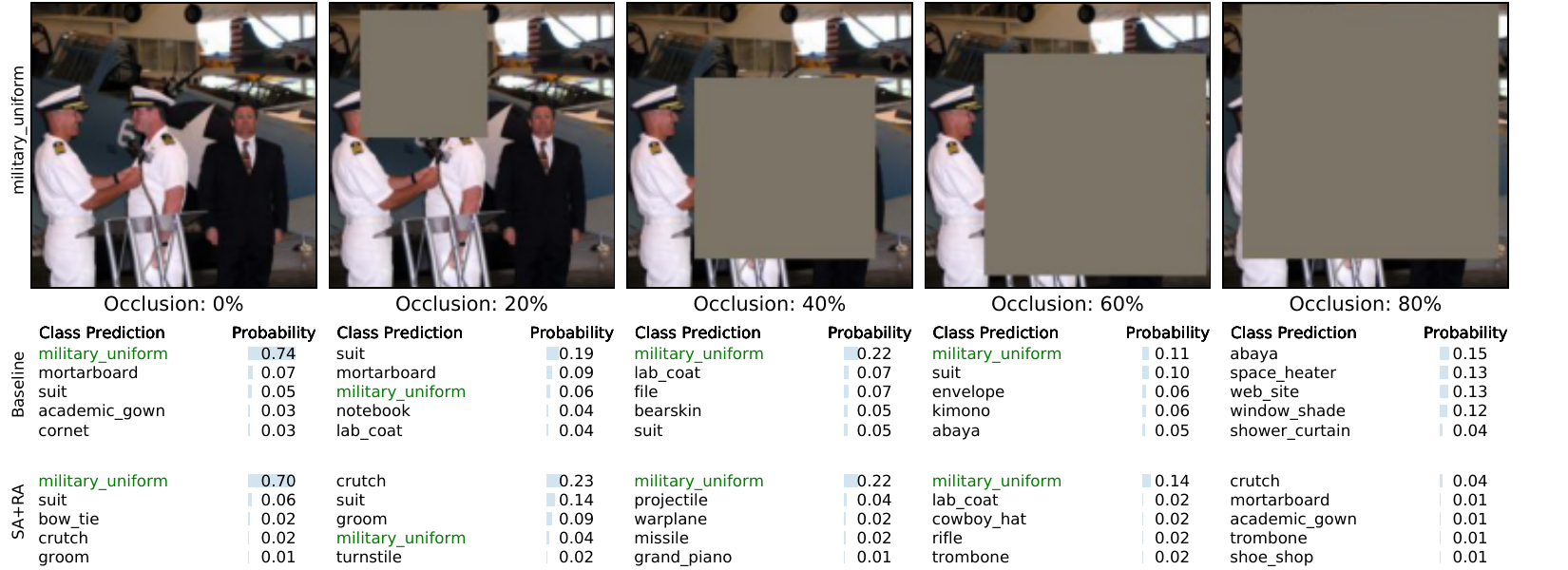}
    \caption{Examples of occluded ImageNet validation images and model predictions of ResNet-50.}
    \label{fig:occlusion_example_additional}
\end{figure*}
\clearpage

\subsection{Self-Supervised Cifar-10/100}\label{sec:ssl_appendix}
Self-supervised SimSiam experiments are run on a single Nvidia A6000 GPU. We follow the standard two-step training recipe~\cite{simsiam}. 1) We first train the Siamese network in a self-supervised manner to learn visual features for 500 epochs with a cosine decay schedule and a batch size of 512. We apply Soft Augmentation only during this step. 2) The linear layer is then tuned with ground-truth labels for 100 epochs with an initial learning rate of 10 and $10 \times$ decay at epochs 60 and 80. Following~\cite{simsiam}, we set $scale_{min} =0.2, scale_{max} = 1.0, ratio_{min} = 3/4, ratio_{max} = 4/3$. Since down-weighting training samples in a batch effectively reduces learning rate and SimSiam is sensitive to it, we normalized the weight in a batch so that the mean remains 1 and re-tuned the learning rate (Table \ref{tab:ssl_appendix}).

\begin{table}[h]
    \centering
        \caption{Soft Augmentation improve self supervised learning with SimSiam. Mean $\pm$ standard error of top-1 validation errors of three runs of ResNet-18 are reported.}
    \resizebox{0.48\textwidth}{!}{
        \begin{tabular}{c c c c c c c}
        \toprule
         Task &  lr & baseline & SA\#1 & $\Delta$ \#1 & SA\#2 & $\Delta$ \#2 \\
        \midrule
         \multirow{6}{*}{Cifar100} & 0.1 & $39.50_{\pm 0.13}$ & $40.21_{\pm 0.03}$  & $+0.71$ & $37.39_{\pm 0.06}$  & $-2.11$\\
          
          & 0.2 & $37.64_{\pm 0.06}$ & $\bm{36.61}_{\pm 0.05}$ & $-1.03$ & $39.20_{\pm 0.42}$  & $+1.56$\\
          
          
          & 0.4 & $40.28_{\pm 2.49}$ & $37.68_{\pm 0.06}$ & $-2.60$ & Diverged & -\\
          
          & 0.5 & $43.26_{\pm 3.03}$ & $41.94_{\pm 0.04}$ & $-1.32$ & Diverged & -\\
          
          & 0.8 & $78.88_{\pm 9.05}$ & $55.44_{\pm 4.15}$ & $-23.44$ & Diverged & -\\
        \midrule  
         \multirow{1}{*}{Cifar10} & 0.2 & $9.87_{\pm 0.03}$ & $\bm{9.31_{\pm 0.01}}$ & $-0.56$  & - & -\\
        \bottomrule
    \end{tabular}
    }
    \label{tab:ssl_appendix}
\end{table}

\begin{table}[h]
    \centering
        \caption{SimSiam k tuning on Cifar-100 (single run)}
    \begin{tabular}{c c c}
        \toprule
         learning rate & k &  Top-1 Error \\
         \midrule
         \multirow{4}{*}{0.2} & 1 & 37.78 \\
         & 2 & 37.27 \\
         & 3 & 36.34 \\
         & 4 & \bm{$36.31$} \\ 
        \bottomrule
    \end{tabular}

    \label{tab:ssl_appendix_k}
\end{table}

\section{Effects of Target Smoothing and Loss Reweighting on Loss Functions}

Consider the KL divergence loss of a single learning sample with a one-hot ground truth vector $\bm{y^{true}}$, and the softmax prediction vector of a model is denoted by $\bm{y^{pred}}$:
\begin{align}
    \label{eq:loss}
    &L(\bm{y^{pred}},\bm{y^{true}}) &= &w * D_{KL}(\bm{y^{true}}||\bm{y^{pred}})\notag \\
    & &= &w * \sum_{n=1}^{N}y^{true}_n * log(\frac{y^{true}_n}{y^{pred}_n}), 
\end{align}

let $n*$ be the ground truth class of an $N$-class classification task, Equation \ref{eq:loss} can be re-written as:

\begin{multline}
    \label{eq:loss_rw}
    L(\bm{y^{pred}},\bm{y^{true}}) = -w * y^{true}_{n*} * log(y^{pred}_{n*}) \\
    + w * \left[ y^{true}_{n*} * log(y^{true}_{n*}) + \sum_{n\neq n*}^{}y^{true}_n * log(\frac{y^{true}_n}{y^{pred}_n})\right] .
\end{multline}

In the case of hard one-hot ground truth target where $y^{true}_{n*} = 1$ and $y^{true}_{n} = 0, n \neq n*$, with the default weight $w=1$ it degenerates to cross entropy loss:

\begin{equation}\label{eq:ce_loss}
    L(\bm{y^{pred}},\bm{y^{true}}) = -log(y^{pred}_{n*}),
\end{equation}

Now we apply label smoothing style softening to the one-hot target $y^{true}$ so that $y^{true}_{n*} = p$ and $y^{true}_{n} = (1-p)/(N-1) = q, n \neq n*$:

\begin{multline}\label{eq:loss_st}
    L(\bm{y^{pred}},\bm{y^{true}}) = -p * log(y^{pred}_{n*}) \\
    + \left[p * log(p) + \sum_{n\neq n*}^{} q * log(\frac{q}{y^{pred}_n})\right]. 
\end{multline}

If $q$ is not distributed, and $y^{true}_{n} = 0, n \neq n*$ (This configuration does not correspond to any of our experiments):

\begin{equation}\label{eq:loss_no_q}
    L(\bm{y^{pred}},\bm{y^{true}}) = -p*log(y^{pred}_{n*}) + p * log(p),
\end{equation}

When only weight $w$ is softened to $w=p$:

\begin{equation}\label{eq:loss_sw}
    L(\bm{y^{pred}},\bm{y^{true}}) = -p * log(y^{pred}_{n*}).
\end{equation}

Note that $p$ is not a function of model weights, so when we take the derivative w.r.t. model weights to compute gradient, Equations \ref{eq:loss_no_q} and \ref{eq:loss_sw} yield the same gradient.

When both the one-hot label and weight are softened with $p$:

\begin{multline}
    \label{eq:loss_sb}
    L(\bm{y^{pred}},\bm{y^{true}}) = -p^2*log(y^{pred}_{n*}) 
    \\
    + p * \left[ p*log(p) + \sum_{n\neq n*}^{} q * log(\frac{q}{y^{pred}_n})\right].
\end{multline}

 The three types of softening in Section \ref{sec:exp} are unique as suggested by Equations \ref{eq:loss_st}, \ref{eq:loss_sw}, and \ref{eq:loss_sb}.

\end{document}